%% file: aaai22.tex
\def\year{2022}\relax
\documentclass[letterpaper]{article} 
\usepackage{aaai22}  
\usepackage{times}  
\usepackage{helvet}  
\usepackage{courier}  
\usepackage[hyphens]{url}  
\usepackage{graphicx} 
\usepackage{natbib}  
\usepackage{caption} 
\DeclareCaptionStyle{ruled}{labelfont=normalfont,labelsep=colon,strut=off} 
\usepackage{algorithm}
\usepackage{algorithmic}
\usepackage{microtype}

\usepackage{newfloat}
\usepackage{listings}
\floatstyle{ruled}
\newfloat{listing}{tb}{lst}{}
\floatname{listing}{Listing}
\usepackage{amsfonts}
\usepackage{amsmath}
\usepackage{booktabs}
\usepackage{graphicx}
\usepackage[table]{xcolor}         

\newcount\Comments
\newcount\ResolvedComments
\definecolor{darkgreen}{rgb}{0,0.5,0}
\newcommand{\kibitz}[2]{\ifnum\Comments=1\textcolor{#1}{#2}\fi}

\newcommand{\resolved}[1] {\ifnum\ResolvedComments=1\textcolor{blue}{[#1]}\fi}

\title{The Spotlight: A General Method for Discovering Systematic Errors in Deep Learning Models}
\author {
    Greg d'Eon\textsuperscript{\rm 1},
    Jason d'Eon\textsuperscript{\rm 2,3},
    James R. Wright\textsuperscript{\rm 4},
    Kevin Leyton-Brown\textsuperscript{\rm 1}
}
\affiliations {
    \textsuperscript{\rm 1}University of British Columbia
    \textsuperscript{\rm 2}Vector Institute
    \textsuperscript{\rm 3}Dalhousie University
    \textsuperscript{\rm 4}University of Alberta \\
    gregdeon@cs.ubc.ca,
    jndeon@dal.ca,
    james.wright@ualberta.ca,
    kevinlb@cs.ubc.ca
}

\begin{document}

\maketitle

\begin{abstract}
\input{sections/00-abstract}
\end{abstract}

\section{Introduction}
\input{sections/01-intro}

\section{Related Work}
\input{sections/02-related_work}

\section{The Spotlight}
\input{sections/03-method}

\section{Experiments}
\input{sections/04-experiments}

\input{sections/05-clustering_comparison}

\section{Future Directions}
\input{sections/06-discussion}

\section{Conclusions}
\input{sections/07-conclusion}


\bibliography{aaai22.bib}

\clearpage
\appendix
\section{Dataset Details}
\input{sections/appendix-datasets}

\section{Additional Results}
\input{sections/appendix-results}

\end{document}

%% file: sections/00-abstract.tex
Supervised learning models often make systematic errors on rare subsets of the data.
When these subsets correspond to explicit labels in the data (e.g., gender, race) such poor performance can be identified straightforwardly. This paper introduces a method for discovering systematic errors that do not correspond to such explicitly labelled subgroups. 
The key idea is that similar inputs tend to have similar representations in the final hidden layer of a neural network. We leverage this structure by ``shining a spotlight'' on this representation space to find contiguous regions where the model performs poorly. 
We show that the spotlight surfaces semantically meaningful areas of weakness in a wide variety of existing models spanning computer vision, NLP, and recommender systems.

%% file: sections/01-intro.tex
Despite their superhuman performance on an ever-growing variety of problems, deep learning models that perform well on average often make systematic errors, performing poorly on semantically coherent subsets of the data.
A landmark example is the Gender Shades study~\citep{Buolamwini2018}, which showed that vision models for gender recognition tend to exhibit abnormally high error rates when presented with images of black women.
AI systems have also been shown to perform poorly for marginalized groups in 
object recognition~\citep{De2019}, speech recognition~\citep{Koenecke2020}, mortality prediction~\citep{Chen2018}, and recruiting tools~\cite{Chen2018}.
Other systematic errors can be harder for practitioners to anticipate in advance. 
Medical imaging classifiers can be sensitive to changes in the imaging hardware~\citep{D2020}; essay scoring software can give high scores to long, poorly-written essays~\citep{Perelman2014}; and visual question-answering systems can fail when  questions are rephrased~\citep{Shah2019}.

Recognizing and mitigating such systematic errors is critical to avoid designing systems that will exhibit discriminatory or unreliable behaviour. Identifying edge cases where systems fail can also be useful for experts seeking to improve their systems \cite{Cai2019}. 
These issues have led the community to develop better tools for testing model performance, clearer standards for reporting model biases, and a plethora of methods for training more equitable or robust models.
However, these methods require practitioners to recognize and label well-defined groups in their dataset ahead of time, 
necessarily overlooking semantically related sets of inputs that the practitioner failed to identify in advance.
While practitioners should certainly continue to assess model performance on sensitive subpopulations, it is extremely difficult to anticipate all of the sorts of inputs upon which models might systematically fail: e.g., vision models could perform poorly on a particular age group, pose, background, lighting condition, etc.

In this work, we introduce \emph{the spotlight}, a method for finding systematic errors in deep learning models even when the common feature linking these errors was not anticipated by the practitioner and hence was not surfaced in an explicit label.
The key idea is that similar inputs tend to have similar representations in the final hidden layer of a neural network.
We leverage this similarity by ``shining a spotlight'' on this representation space, searching for contiguous regions in which the model performs poorly.
We show that the spotlight surfaces semantically meaningful areas of weakness in a surprisingly wide variety of otherwise dissimilar models and datasets, including image classifiers, language models, and recommender systems.

Rather than replacing existing methods for measuring biases in trained models, we hope that practitioners will add the spotlight to their model-auditing pipelines to identify failure modes that do not correspond to known labels. After such problems are identified by the spotlight, developers  can then mitigate them by augmenting their datasets, adjusting their model architectures, or leveraging robust optimization methods.
We provide an open-source implementation of the spotlight at [URL withheld for anonymous review].%

%% file: sections/02-related_work.tex
\paragraph{Systematic errors on known groups.}
A standard approach for auditing a machine learning model is to create a dataset partitioned by group information, (e.g., demographics, lighting conditions, hospital ID, \dots), and to check whether the model exhibits poor performance on any of these groups.
With one or two particularly sensitive group variables, it is straightforward to check the model's performance on each group; 
e.g., the NLP community advocates for including such disaggregated evaluations in model cards~\cite{Mitchell2019}.
When there are more group variables, inducing exponentially many intersectional groups, there are a variety of computational methods to efficiently identify groups for which performance is poor~\cite{Chung2019, Li2021, Pastor2021} and model dashboards for interactively exploring groups~\citep{Cabrera2019, Ahn2019, Wexler2019}.
  
After identifying a systematic problem with a model, there is a substantial literature proposing methods for making repairs.
Fair machine learning methods can incorporate group information into shallow models, requiring that the model perform similarly on each group \citep[see, e.g.,][for a review]{Corbett2018} or evaluating the model on its worst group~\cite{Martinez2020}. There also exist generalizations to support many intersectional groups~\cite{Kim2018}.
Two recent robust training methods exist to repair deep models that exhibit such biases. 
First, distributionally robust language modelling~\cite{Oren2019} allows an adversary to change the distribution over groups during training, requiring the model to do well on each group.
Second, invariant learning~\cite{Arjovsky2019} requires the model to learn a representation of the data that induces the same classifier for each group, protecting against spurious correlations.

The spotlight differs from these approaches in that it aims to identify systematic failure modes that are not described by existing group labels.
Our approach is thus complementary to those just surveyed: after using the spotlight to uncover a new failure mode, practitioners can augment their dataset with appropriate labels and turn to these existing methods to monitor or repair their model's performance. 

\paragraph{Systematic errors on unknown groups.}
One approach to avoiding systematic errors without group information is to train a model that performs well across the entire dataset; such a model is guaranteed to achieve good performance on any semantically meaningful subset of the data.
For example, distributionally robust optimization (DRO) methods can work without group information by allowing an adversary to select from a broader set of dataset reweightings~\cite{Hashimoto2018,Duchi2020,Lahoti2020}; invariant learning algorithms can infer groups during training~\cite{Creager2020}.
While these methods sometimes leverage information from a trained model's representation space, they differ from the spotlight by attempting to mitigate systematic errors within the training loop.
In contrast, the spotlight aims to surface any biases that remain after model training to a human expert.

The existing method most similar to our own approach is GEORGE~\citep{Sohoni2020}, a method for applying DRO without group labels. 
GEORGE infers  ``subclasses'' within the dataset by clustering points within a trained neural network's representation space, then allows an adversary to modify the distribution over subclasses.
While Sohoni et~al.\ focused on training robust models, they observe that these clusters tend to correspond to semantically meaningful subsets of the data (for instance, images of birds on land vs. on water). 
They also observe that their reliance upon a superlinear-time clustering method limits its applicability to large datasets.
The spotlight exploits the same underlying insight from GEORGE that semantic similarity is reflected by proximity in the embedding space. However, the spotlight focuses on auditing models rather than robust training; dramatically lowers computational cost to linear time; avoids partitioning the entire embedding space, searching only for contiguous, high-loss regions; and can identify issues that involve examples from multiple classes.

A final notable method is Errudite~\cite{Wu2019}, an interactive system for analyzing errors made by NLP models.
It allows a user to query a subset of their dataset using a domain-specific language, reporting the model's performance on this query set and proposing related queries to help the user dig deeper into their model.
This interactive query system can help developers discover and confirm systematic issues in their models without the need for pre-existing group labels.
However, Errudite is carefully designed to suit models for NLP tasks; the spotlight is domain-agnostic, applying to a broad family of deep models on many different domains.

%% file: sections/03-method.tex

\begin{figure}
    \centering
    \includegraphics[scale=0.9]{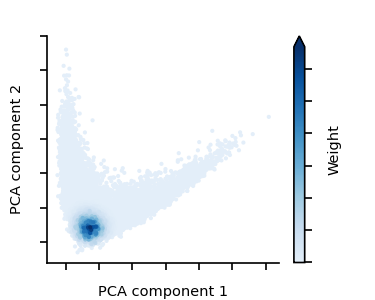}
    \caption{An example of a spotlight in a model's representation space. }
    \label{fig:example-spotlight}
\end{figure}

In order to identify systematic errors, we examine reweightings of the dataset that emphasize poor performance of the model.
The reweightings are computed by applying a parameterized kernel function to the final representation space of the model, concentrated on a contiguous region, without only focusing on a few points.
We refer to such a reweighting as a \emph{spotlight} (Figure~\ref{fig:example-spotlight}).

Formally, we suppose that we have $N$ data points with representations $x_1, \dots, x_N \in \mathbb{R}^d$ and losses $\ell_1, \dots, \ell_N \in \mathbb{R}$.
We compute weights as $k_i = \max(1 - \tau (x - \mu)^2, 0)$ where $\mu \in \mathbb{R}^d$ and $\tau \in \mathbb{R}$ are the center and precision of the spotlight, respectively.
Note that $k_i$ has a maximum of $1$ when $x_i = \mu$ and a minimum of $0$ when $x_i$ is sufficiently far from $\mu$.
Then, our objective is
\begin{align}
    \max_{\mu, \tau} \quad & \sum_i{\left( \frac{k_i}{\sum_j{k_j}} \right) \ell_i }\\
    \textrm{s.t.} \quad & \sum_i{k_i} \geq S,
\end{align}
for some choice of the hyperparameter $S$. 
We interpret $S$ as the ``spotlight size,'' as this setting ensures a lower bound on the total weight that the spotlight assigns across the dataset. 

To make optimization tractable, we replace the hard constraint $\sum_i k_i \ge S$ with the penalty term
\begin{equation}
    b(k) = C \cdot \max \left( \frac{(\sum_i k_i - (S + w))^2}{w^2}, 0 \right)
\end{equation}
and optimize the unconstrained objective
\begin{equation}
    \label{eqn:spotlight-objective}
    \sum_i{\left( \frac{k_i}{\sum_j{k_j}} \right) \ell_i } - b(k).
\end{equation}
This penalty acts as a ``barrier,'' penalizing reweightings that include less than $S+w$ points.
We fix $C$ to be large relative to typical losses in the dataset (i.e. $C = 1$ for binary classification; $C = 10$ for problems with thousands of classes) and gradually decrease $w$ while optimizing.

To optimize Equation~\ref{eqn:spotlight-objective}, we begin with a large, diffuse spotlight containing the entire dataset (in practice, initializing to $\mu = 0$ and $\tau = 10^{-4}$).
Then, we run the Adam optimizer for 5000 steps with an adaptive learning rate, halving the learning rate each time the objective reaches a plateau, and we gradually shrink the width of the barrier geometrically from $w = N-S$ to $w = 0.05S$.

We additionally developed a method of optimizing multiple distinct spotlights on the same dataset without changing any hyperparameters. This method iteratively subtracts spotlight weight from its internal accounting of each example's loss and then finds another spotlight. More formally, using the weights provided by spotlight, $k_i$, we update the losses as $\ell_i^\prime := \left( 1 - k_i / \max_j k_j \right) \ell_i$;
we then perform the same optimization as above and repeat as many times as desired. In all of our experiments, we present multiple spotlights obtained in this way.

Since the sum of the weights, $k_i$, has an upper bound of $N$, we considered spotlight sizes ranging from $S = 0.001N$ (0.1\%) to $S = 0.1N$ (10\%).
We found that spotlights on the smaller end of this spectrum were too selective to observe any cohesion, whereas the largest spotlights were too inclusive.
Taking this into account, we settled on a spotlight size of $2\%$ for vision models where images can be simply be scanned for cohesion, and a spotlight size of $5\%$ for non-vision models, where we found it necessary to describe spotlights using summary statistics.





%% file: sections/04-experiments.tex
\label{sec:experiments}
The spotlight is relatively model-agnostic: it only requires knowledge of the final layer representations and the losses for each input.
We demonstrate this flexibility by using the spotlight to evaluate a broad range of pre-trained models from the literature, spanning image classification (faces; objects; x-rays), NLP (sentiment analysis; question answering), and recommender systems (movies). 
In each case, we show that the spotlight uncovers systematic issues that would have otherwise required group labels to uncover.
Full results can be found for every dataset we tested (most in the appendix), showing the method's surprising reliability.
Experiments were run a single NVIDIA Tesla V100 GPU, with which each spotlight presented in this section took under 1 minute to optimize, emphasizing the computational tractability of our approach even on very large datasets.

We also compare these spotlights with the clustering stage of GEORGE's pipeline.
In particular, we use the publicly available implementation of GEORGE\footnote{\url{https://github.com/HazyResearch/hidden-stratification}}, which separately clusters the examples from each class, automatically selecting the number of clusters using a Silhouette-based heuristic.
Wherever it is feasible, we identify the three highest-loss clusters, summarize the examples in each of these clusters, and compare them to the content within the spotlights.
We show that the content of the spotlights often differ from these clusters, finding more granular problem areas within a class or systematic errors that span across multiple classes.

\subsection{FairFace}
\begin{figure*}
    \centering
    Random sample: \\
    \includegraphics[trim={0 1in 0 0},clip]{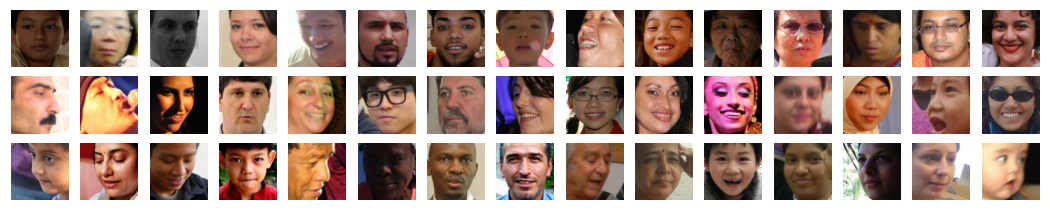}
    
    Highest losses: a diffuse set \\
    \includegraphics[trim={0 0.7in 0 0},clip]{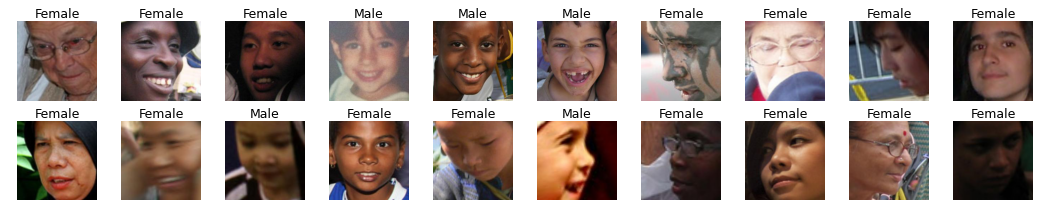}
    
    Spotlight 1: side profile views/poor framing \\
    \includegraphics[trim={0 0.7in 0 0},clip]{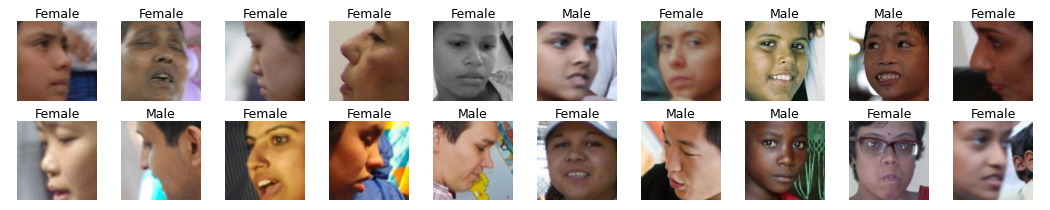}
    
    Spotlight 2: Asian children \\
    \includegraphics[trim={0 0.7in 0 0},clip]{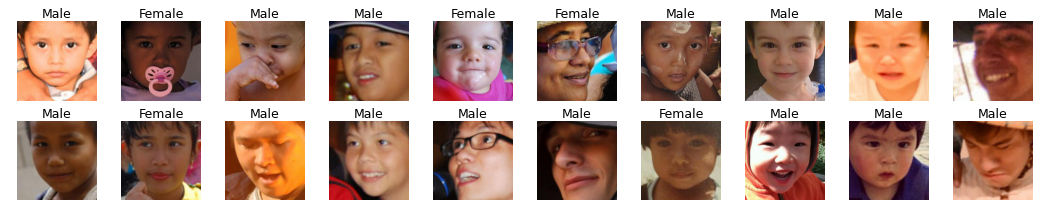}
    
    Spotlight 3: obscured/shadowed faces \\
    \includegraphics[trim={0 0.7in 0 0},clip]{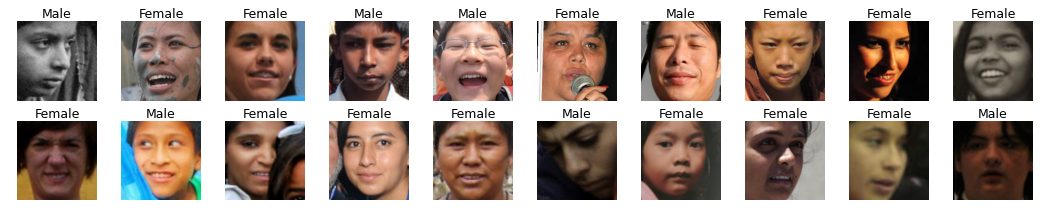}
    
    \caption{Spotlights on FairFace validation set. Image captions list true label.}
    \label{fig:fairface}
\end{figure*}

We first study FairFace~\citep{Karkkainen2021}, a collection of 100,000 face images annotated with crowd-sourced labels about the perceived age, race, and gender of each image.
FairFace is notable for being approximately balanced across 7 races and 2 genders.
In particular, we trained a model to predict the perceived gender label as a proxy for the gender prediction systems studied in prior work~\citep{Buolamwini2018}.
Our model was a ResNet-18, trained using Adam with cross-entropy loss and a learning rate of 3e-4; we stopped training after 2 epochs when we found that the validation loss began increasing.
We ran the spotlight on the validation set, using the final 512-dimensional hidden layer for the representation space.

The spotlights and highest-loss examples are shown in Figures~\ref{fig:fairface} and \ref{fig:fairface-extra}.
We found that each of the spotlights discovered a strikingly different set of faces, each representing a problem area for the model.
The first shows a set of profile (i.e., side) views, where it is difficult to see many of the facial features; the second consists mostly of young children whose genders are relatively harder to discern; the third shows some faces that are shadowed or obscured.
The fourth and fifth spotlights consist of black faces in poor lighting and Asian faces, respectively.
(These demographic disparities are confirmed in Figure~\ref{fig:fairface_demographics}, which shows the distribution of ages and races on each spotlight.)
Overall, our spotlights identified both age and racial groups that the model performs poorly on---without access to these demographic labels---and semantically meaningful groups for which labels did not exist.
In comparison, the high-loss images are an unstructured set of examples that include occluded faces, poor lighting, blurry shots, and out-of-frame faces.

GEORGE identified a total of six clusters.
A random sample of the images from the three clusters with the highest average losses are shown in Figure~\ref{fig:fairface-extra}.
Notably, all of the images in each cluster have the same label, but share little in common beyond their labels.
Like the high-loss images, each of GEORGE's clusters display a combination of camera angles, lighting conditions, and occlusions that make the images difficult to classify (as opposed to the spotlights, which identify a single semantically meaningful issue).

\subsection{ImageNet}
For a second vision dataset, we study the pre-trained ResNet-18 model from the PyTorch model zoo~\citep{Zoo}, running the spotlight on the 50,000 image validation set.
As in FairFace, we used the final 512-dimensional hidden layer of the model as the representation space.

\begin{figure*}
    \centering
    Random sample: \\
    \includegraphics[trim={0 1in 0 0},clip]{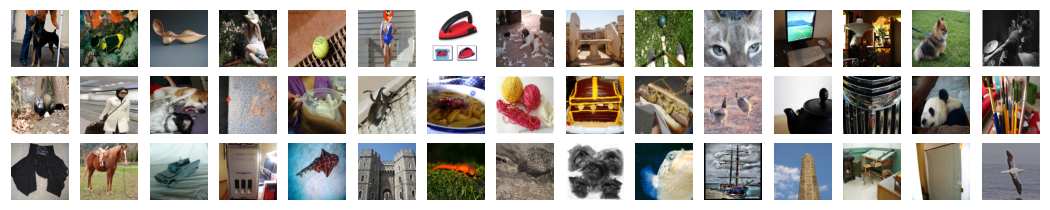}
    
    Highest losses: a diffuse set \\
    \includegraphics[trim={0 0.7in 0 0},clip]{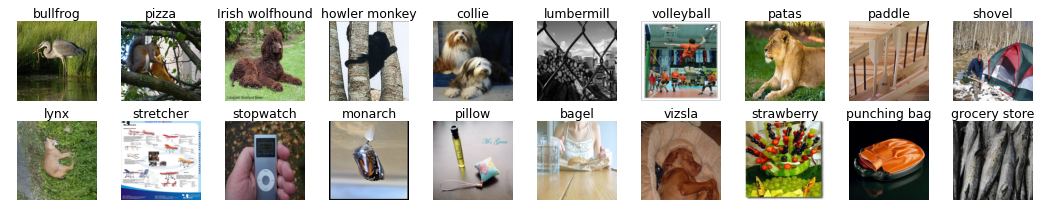}
    
    Spotlight 1: people working
    \includegraphics[trim={0 0.7in 0 0},clip]{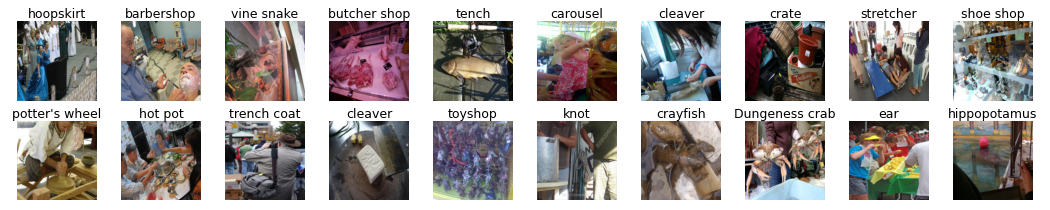}
    
    Spotlight 2: tools and toys
    \includegraphics[trim={0 0.7in 0 0},clip]{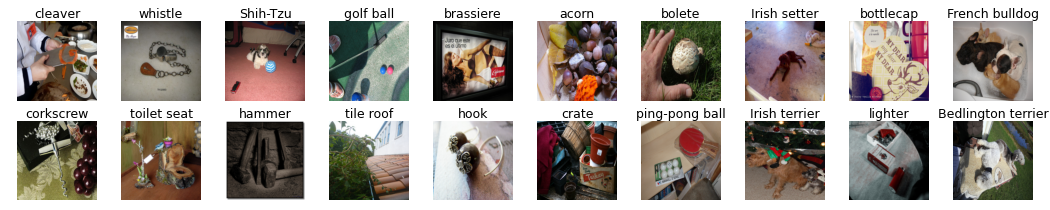}
    
    Spotlight 3: animals in foliage
    \includegraphics[trim={0 0.7in 0 0},clip]{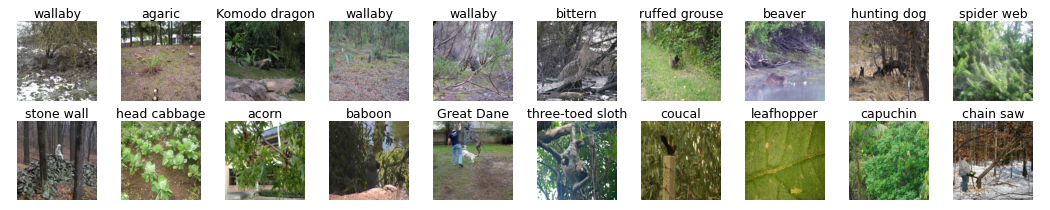}
    
    \caption{Spotlights on ImageNet validation set. Image captions list true label.}
    \label{fig:imagenet}
\end{figure*}

Our results are shown in Figures~\ref{fig:imagenet} and~\ref{fig:imagenet-extra}.
Each spotlight found a set of images that have a clear ``super-class'', but are difficult to classify beyond this super-class.
The first spotlight contains a variety of images of people working, where it is difficult to tell whether the label should be about the person in the image, the task they're performing, or another object; the second shows a variety of tools; the third shows a variety of green plants, where there is often an animal hiding in the image; the fourth identifies some food and people posing; and the fifth shows a variety of dogs.
In contrast, the high-loss images appear to have little structure, with many of them having unexpected labels, such as ``pizza'' for an image of a squirrel in a tree holding a piece of pizza.

We did not run GEORGE on this model, as the clustering algorithm failed when attempting to split the 50 images in each class into even smaller clusters.
However, GEORGE's clusters would be much different from the spotlights, as its ``subclass'' clusters can only de-aggregate existing classes, whereas the spotlight tended to reveal semantically similar groups of classes---``superclasses''---that the model had trouble distinguishing. 

\subsection{Sentiment analysis: Amazon reviews}
Next, we turn to the Amazon polarity dataset~\cite{Zhang2015}, a collection of 4 million plain-text Amazon reviews labelled as ``positive'' (4-5 stars) or ``negative'' (1-2 stars).
We used a popular pre-trained checkpoint of a DistilBERT model from Huggingface~\citep{SST2}, which was fine-tuned on SST-2. 
We ran the spotlight on a sample of 20,000 reviews from the validation set, using the final 768-dimensional hidden layer as the representation space.

We found it more difficult to spot patterns in the spotlights on this dataset by simply reading the highest-weight reviews, so we instead summarized each spotlight by identifying the tokens that appeared most frequently in the spotlight distributions, relative to their frequencies in the validation set.
These results are shown in Figure~\ref{fig:amazon}.
Remarkably, the first spotlight surfaced reviews that were written in Spanish, which the model consistently classifies as negative: it was only trained on English sentences, and its tokenizer appears to work poorly on Spanish sentences.
The second spotlight highlighted long-winded reviews of novels, which the model has difficulty parsing; the third found reviews that mention aspects of customer service, such as product returns, which the model classifies as extremely negative. 

The highest-loss reviews in the dataset are quite different, consisting almost entirely of mislabelled reviews.
For example, one review reads ``The background music is not clear and the CD was a waste of money. One star is too high.'', but has a 4-5 star rating; dozens of high-loss outliers follow this pattern, where the rating clearly contradicts the review text.
We note that this type of label noise would pose a problem for many robust optimization methods, which could insist that the model learn to memorize these outliers rather than focusing on other important portions of the dataset.

GEORGE found a total of 11 clusters; the three with the highest losses are also summarized in Figure~\ref{fig:amazon}.
The first cluster is similar to the second spotlight, containing many negative and wordy reviews for novels and movies that are misleading or difficult to parse.
The second consists entirely of positive reviews, including many written in Spanish. 
The third is small, only containing 0.5\% of the dataset, and we found it difficult to summarize.
Overall, several of the features found by the spotlights also appear in GEORGE's clusters; a more in-depth analysis would help to understand how the two methods differ on this model.

\begin{figure*}
    \centering
    \input{tables/amazon-results}
    \caption{Spotlight on Amazon reviews.}
    \label{fig:amazon}
\end{figure*}

\subsection{MovieLens 100k}
We investigated a third domain, recommender systems. Specifically, we considered the MovieLens 100k dataset~\citep{Movielens}, a collection of 100,000 movie reviews of 1,000 movies from 1,700 different users.
It also includes basic information about each movie (titles, release dates, genres) and user (age, gender, occupation), which we use during the analysis, but did not make available to the model.
For our model, we used a deep factorized autoencoder~\citep{Hartford2018}, using the final 600-dimensional hidden layer for our representation space.

The highest-weight movies in each spotlight are shown in Figure~\ref{fig:movielens-spotlights}.
The first spotlight mostly identifies 3-4 star action and adventure films rated by prolific users, where the model is highly uncertain about which review they will give.
The second finds reviews of highly rated drama films from a small group of users with little reviewing history.
The third shows unpopular action and comedy films, where the model is nonetheless optimistic about the rating. 
In comparison, the highest-loss predictions consist mostly of 1-star ratings on movies with high average scores.

GEORGE identified 21 clusters; we show the highest-loss predictions from three in Figure~\ref{fig:movielens-george}.
The first two consist of a variety of 1- or 2-star ratings respectively, where the model confidently makes 4- or 5-star predictions for both categories.
Both clusters have similar genre distributions to the entire dataset.
The third cluster instead contains many 5-star ratings on comedy and drama films where the model is skeptical about these high ratings.
The GEORGE clusters differ from the spotlights, which tend to have more consistent movies or genres, but less consistent ratings.

\subsection{Additional Datasets: SQuAD and Chest X-rays}
We ran the spotlight on two additional datasets: SQuAD, an NLP question-answering benchmark; and a chest x-ray image dataset.
Our results here were more ambiguous, but we describe these experiments regardless to emphasize the spotlight's generality and to reassure the reader that we have presented all of our findings rather than cherry-picking favourable results. 
Full details can be found in the appendix. 

\paragraph{SQuAD.}
The Stanford question answering dataset (SQuAD)~\citep{Rajpurkar2016} is a benchmark of question-answer pairs constructed from the content of 536 Wikipedia articles.
We analyzed a pre-trained DistilBERT model~\citep{SQuAD} fine-tuned on this dataset, running spotlights on the test set.
We excluded long examples where the sum of the context and answer sequence lengths was greater than 384, leaving 10,386 question-answer pairs.
It was unclear which representation to use for the SQuAD spotlights, as the model has one output for each token in the context rather than a single representation for the entire example; we used the last-layer representations for the \texttt{[CLS]} token in each example.

The results are summarized in Figure~\ref{fig:squad}.
The spotlights particularly highlighted questions from the ``packet switching'' and ``civil disobedience'' categories, which were the two categories with the highest loss, despite having no knowledge of these categories; the individual questions with the highest losses identified the latter but not the former.
We found little semantic structure beyond these high-level categories; a richer representation space is likely required to get more insight into this dataset.

\paragraph{Chest x-rays.}
The chest x-ray dataset consists of 6,000 chest x-rays labelled as ``pneumonia'' or ``healthy''~\citep{Kermany2018}.
To summarize, the spotlight identified at least two semantically meaningful failure modes in this domain: images with a text label ``R'' on their sides, and images with very high contrast. 
However, such images were also relatively easy to identify in the set of high-loss inputs, so we were unconvinced that the spotlight offered a decisive benefit in this case. 
Due to our lack of expertise in radiology, we were unable to discover identifying attributes in the spotlighted clusters; it is quite possible that other spotlighted clusters were semantically related in more fundamental ways. 
It is also possible that the small training set size led to a less meaningful embedding space; indeed, this was one of only two datasets for which we were unable to leverage an existing, pre-trained model.


%% file: tables/amazon-results.tex
\begin{tabular}{lr p{3.5in}}
    \toprule
    \textbf{Subset} & \textbf{Avg Length} & \textbf{Frequent words} \\
    \midrule
    High loss & 
    68.8 &
    length, outdated, potter, bubble, contact, cinematography, adjusting, functions, stock, versus \\
    \midrule
    Spotlight 1: Spanish & 
    79.1 &
    que, est, como, y, las, tod, es, la, si, por \\
    \midrule
    Spotlight 2: novels & 
    88.7 &
    super, wearing, job, prefer, bigger, hang, discover, killing, slip, source \\
    \midrule
    Spotlight 3: customer service & 
    80.7 &
    problem, returned, que, hoping, ok, unfortunately, returning, however, las, maybe \\
    \midrule
    GEORGE cluster 1 &
    77.6 &
    ok, moore, fiction, okay, above, potter, thank, cinematography, usa, jean \\
    \midrule
    GEORGE cluster 2 & 
    75.9 &
    que, para, y, est, como, es, tod, las, installation, la \\
    \midrule
    GEORGE cluster 3 &
    87.7 &
    visuals, sugar, investment, score, study, surfer, dimensional, era, dune, scarlet \\
    \bottomrule
\end{tabular}

%% file: sections/05-clustering_comparison.tex
\subsection{Quantitative Evaluation}
\begin{figure*}
    \centering
    \includegraphics{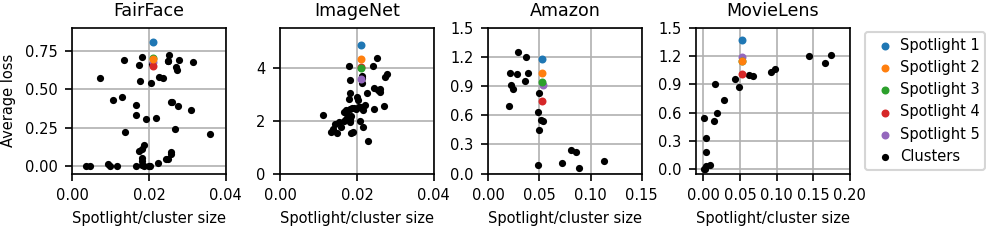}
    \caption{Sizes and average losses for clusters and spotlights. FairFace and ImageNet show spotlights containing 2\% of the dataset and 50 clusters; Amazon reviews and MovieLens show spotlights containing 5\% of the dataset and 20 clusters.}
    \label{fig:cluster_spotlight_losses}
\end{figure*}

Finally, we evaluated whether the spotlights were effective at discovering high-loss regions in each model's representation space.
To do this, we compared to a naive clustering baseline: we fit a Gaussian mixture model to each representation space and compared the spotlight to the cluster with the highest average loss.
We fit 50 clusters on FairFace and ImageNet and 20 clusters on Amazon and MovieLens, ensuring that a typical cluster would be comparable in size to the spotlights.
Then, we compared the size and average loss of each cluster to the spotlights.

The results are shown in Figure~\ref{fig:cluster_spotlight_losses}.
On FairFace, ImageNet, and MovieLens, the first spotlight has a higher loss than any of the clusters, showing that spotlight effectively identifies a high-loss region.
On Amazon, two clusters have a higher average loss than the first spotlight, but they do so by incorporating fewer points, finding a less widespread error.
These results indicate that the spotlights reliably use information about the model's losses to identify a large systematic error, while a naive clustering method tends to split high-loss regions across several clusters or identify smaller failure modes.

%% file: sections/06-discussion.tex
Our methods give rise to various promising directions for future work, many of which we have begun to investigate. This section describes some of these ideas along with our initial findings.

\paragraph{Using the spotlight for adversarial training.}
While this paper advocates for the spotlight as a method for auditing deep learning models, it also gives rise to a natural, adversarial objective that could be optimized during training in the style of the distributionally robust methods surveyed earlier~\cite{Duchi2020,Lahoti2020}. 
That is, model training could iterate between identifying a spotlight distribution, reweighting the input data accordingly, and minimizing loss on this reweighted input.
A model that performed well on this objective would have very balanced performance, distributing inputs with poor performance diffusely across the representation space.
Unfortunately, our preliminary tests suggest that optimizing for this objective is not simple.
With large spotlights (10\% of dataset), we found that this method made little difference, with the model improving more slowly than in regular training; with smaller spotlights (1\%), the model struggled to learn anything, fluctuating wildly in performance between epochs.
We intend to continue investigating approaches for training against this flexible adversary.

\paragraph{Structure in representations.}
An important assumption that the spotlight makes is that nearby points in the representation space will tend to correspond to semantically similar inputs.
While this assumption is empirically supported both by our results and by prior work~\cite{Sohoni2020}, it is  an emergent property of deep learning models, and we do not currently understand this property's sensitivity to details of the architecture and training method.
For instance, does the choice of optimizer (SGD/Adam, weight decay, learning rate, \dots) affect the representation space in a way that interacts with the spotlight? 
Could we instead leverage representations learned by alternative models, such as autoencoders?

\paragraph{Spurious correlations.}
In particular, models that have learned spurious correlations could act quite differently from the models we studied in this work.
For example, consider a model trained on the Waterbirds dataset~\cite{Sagawa2019}, where it is possible to reach high training accuracy by learning to recognize land/water backgrounds instead of correctly identifying land/water birds.
The model's representation would then focus mostly on details of the backgrounds, and spotlights would be unable to substantially change the distribution of bird types.
Investigating a model's representation spaces with tools like the spotlight could help to understand why a model is failing on a particular distribution shift.


%% file: sections/07-conclusion.tex
\label{sec:conclusions}
The spotlight is an automatic and computationally efficient method for surfacing semantically related inputs upon which a deep learning model performs poorly. In experiments, we repeatedly observed that the spotlight was able to discover meaningful groups of problematic inputs across a wide variety of models and datasets, including poorly modelled age groups and races, ImageNet classes that were difficult to distinguish, reviews written in Spanish, difficult question categories, and specific movies with unpredictable reviews.
The spotlight found all of these sets without access to side information such as demographics, topics, or genres.

The spotlight is not a direct solution to the problem of systematic errors in deep learning models, instead fitting 
into a broader feedback loop of developing, auditing, and mitigating models.
The spotlight is useful in the auditing stage of this loop, helping practitioners to discover semantically meaningful areas of weakness that they can then test in more depth and address through changes to their pipeline.
Such a human-in-the-loop discovery process is critical to identify systematic failure modes in deep learning systems and mitigate them before they are able to cause harmful consequences in deployed systems. 

\paragraph{Potential social impacts.} 
The spotlight can be a useful addition to machine learning practitioners' toolboxes, augmenting their existing robustness tests.
However, the spotlight can only show the existence of a systematic error, not prove that a model has none.
It is possible that a practitioner could get a false sense of security if the spotlight turns up no significant issues -- perhaps because their model's biased representation hides an important issue, or because they miss a systematic error on visual inspection.
On balance, the potential to uncover new issues outweighs the risk of believing there are none, especially when the spotlight is used in concert with other fairness or robustness methods.

Additionally, it is conceivable that the spotlight could be used for debugging harmful AI systems, such as surveillance technology, to identify regimes under which these technologies fail and to further improve their efficacy.
This is unavoidable: the spotlight is general enough to work on a wide range of model architectures, including those that might cause negative social impacts. 
Overall, though, we do not see this as a likely use case; the spotlight's main likely effect would be helping practitioners to increase the fairness and robustness of deployed deep learning systems and to gain confidence that their models to not systematically discriminate against coherent subpopulations of users.

%% file: sections/appendix-datasets.tex
\label{sec:datasets}
In Figure~\ref{fig:dataset-details}, we summarize licensing and content considerations for each of the datasets used in this work.

\begin{figure*}
    \centering    
    \begin{tabular}{llll}
         \toprule
         Dataset & License & PII & Offensive content \\
         \midrule
         FairFace       & CC BY 4.0              & none & none \\
         ImageNet       & custom non-commercial  & none & none \\
         Amazon reviews & Apache 2.0             & none & Offensive words in reviews are censored \\
         SQuAD          & CC BY 4.0              & none & none \\
         MovieLens 100K & custom non-commercial  & none & none \\
         Chest x-rays   & CC BY 4.0              & none & none \\
         Adult          & MIT                    & none & none \\
         Wine quality   & MIT                    & none & none \\
         \bottomrule
    \end{tabular}
    \caption{Details for each of the datasets used in this paper.}
    \label{fig:dataset-details}
\end{figure*}

%% file: sections/appendix-results.tex
\begin{figure*}
    \centering
    Spotlight 4: dark skin tones; poor lighting \\
    \includegraphics{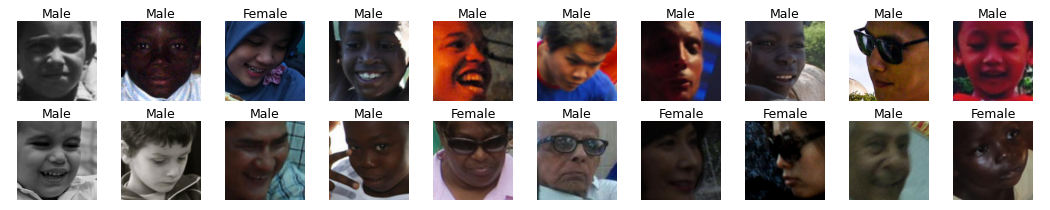}
    
    Spotlight 5: Asian faces \\
    \includegraphics{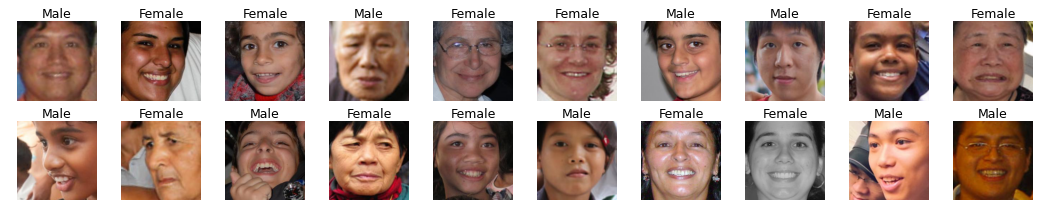}
    
    GEORGE cluster 1 \\
    \includegraphics{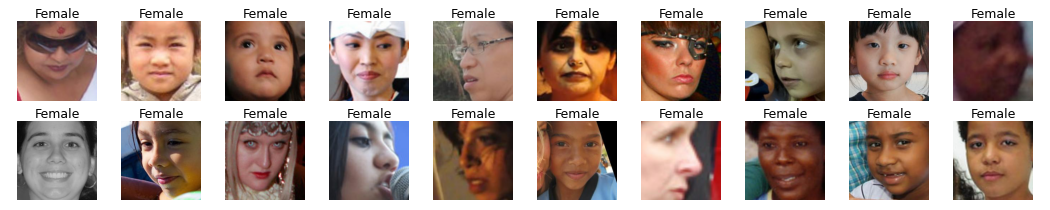}
    
    GEORGE cluster 2 \\
    \includegraphics{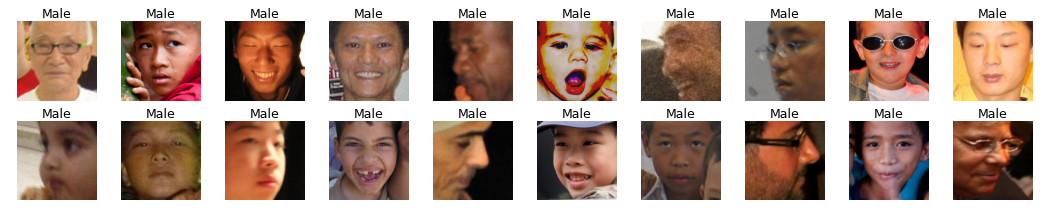}
    
    GEORGE cluster 3 \\
    \includegraphics{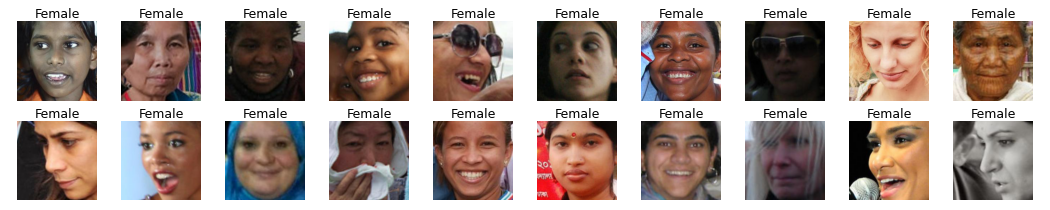}
    
    \caption{Additional spotlights and GEORGE clusters on FairFace.}
    \label{fig:fairface-extra}
\end{figure*}
\begin{figure*}
    \centering
    \includegraphics{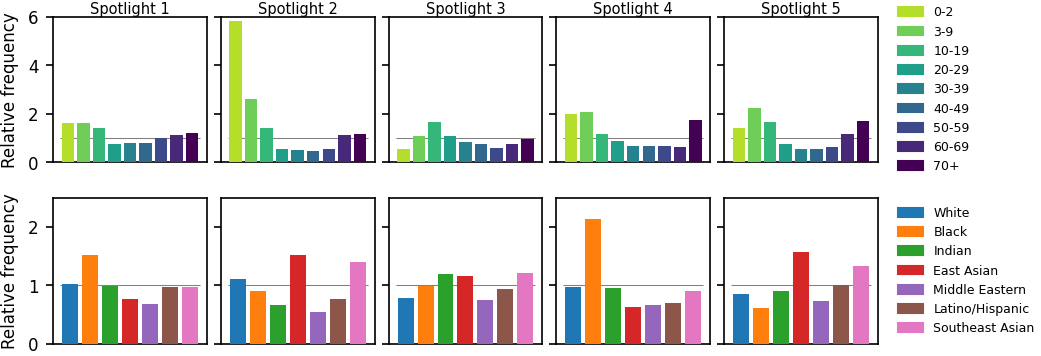}
    \caption{Demographics on FairFace.}
    \label{fig:fairface_demographics}
\end{figure*}
\begin{figure*}
    \centering
    Spotlight 4: food; people posing \\
    \includegraphics{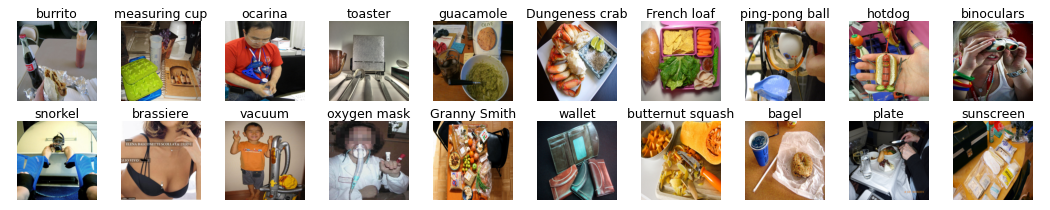}
    
    Spotlight 5: outdoor dogs \\
    \includegraphics{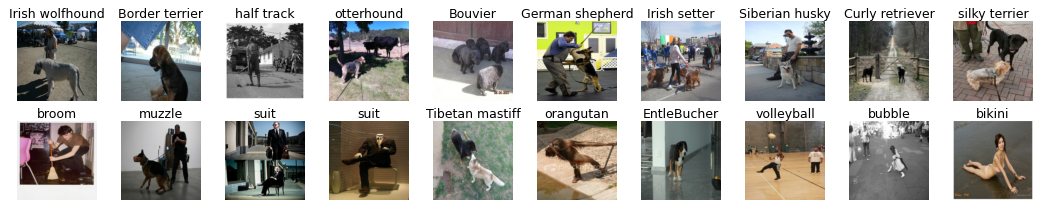}
    
    \caption{Additional spotlights on ImageNet.}
    \label{fig:imagenet-extra}
\end{figure*}
\begin{figure*}
    \centering
    Random sample: \\
    \includegraphics{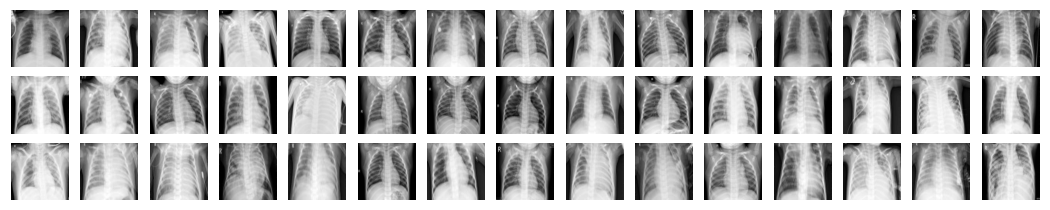}
    
    Highest losses: \\
    \includegraphics{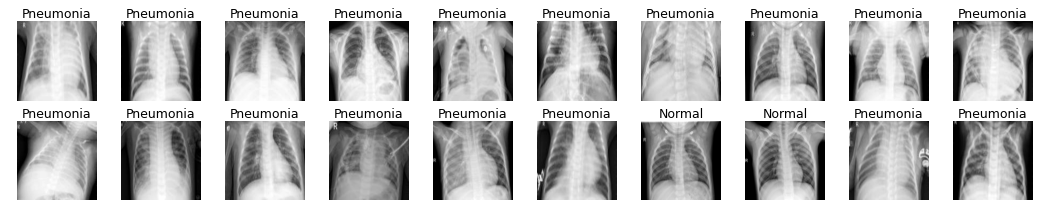}
    
    Spotlight 1: \\
    \includegraphics{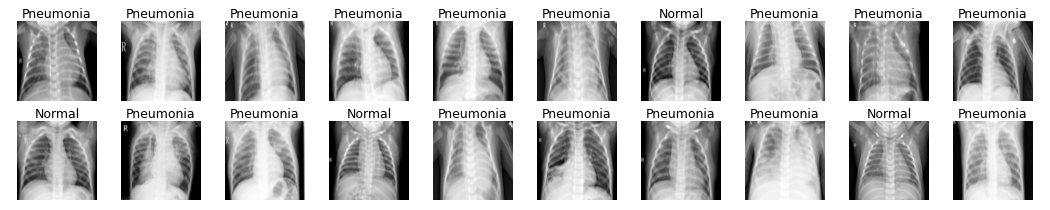}
    
    Spotlight 2: \\
    \includegraphics{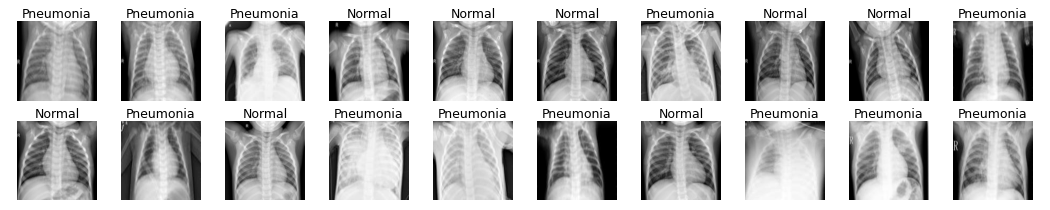}
    
    Spotlight 3: \\
    \includegraphics{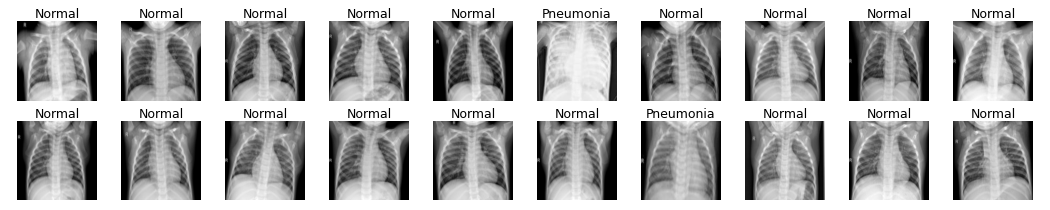}
    
    \caption{Chest xray sample images, high loss images, and spotlights.}
    \label{fig:xray}
\end{figure*}
\begin{figure*}
    \centering
    Spotlight 4: \\
    \includegraphics{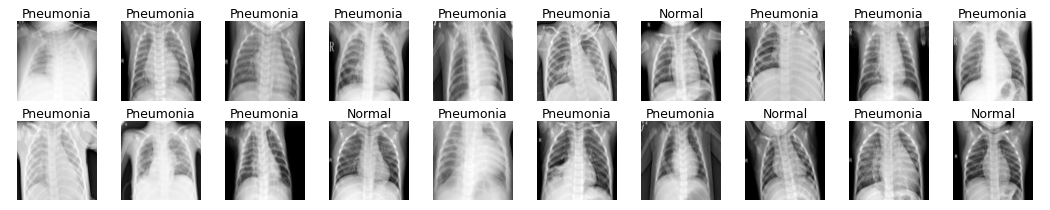}
    
    Spotlight 5: \\
    \includegraphics{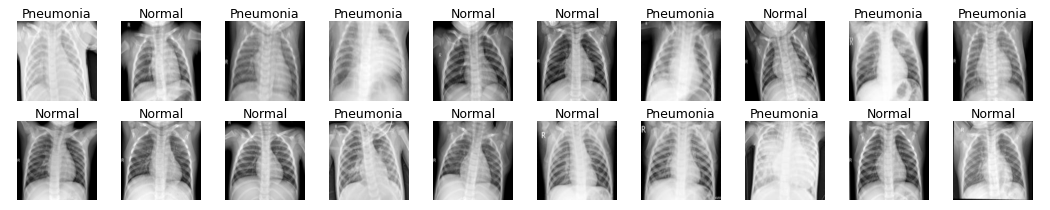}
    
    \caption{Additional chest xray spotlights.}
    \label{fig:xray-extra}
\end{figure*}

\label{sec:extra-results}
In this section, we include additional outputs from the spotlights on the image and MovieLens datasets that were described in the text.
In particular, we include:
\begin{itemize}
    \item FairFace: fourth and fifth spotlights and GEORGE clusters in Figure~\ref{fig:fairface-extra}; demographic info in Figure~\ref{fig:fairface_demographics}
    \item ImageNet: fourth and fifth spotlights in Figure~\ref{fig:imagenet-extra}
    \item MovieLens: high loss ratings in Figure~\ref{fig:movielens-high-loss}; spotlight examples in Figure~\ref{fig:movielens-spotlights}; GEORGE clusters in Figure~\ref{fig:movielens-george}.
    \item Chest x-rays: random sample, high loss examples, and first three spotlights in Figure~\ref{fig:xray}; final two spotlights in Figure~\ref{fig:xray-extra}
    \item SQuAD: common words and topics from each spotlight in Figure~\ref{fig:squad}
\end{itemize}

\clearpage

\begin{figure*}
    \centering
    \input{tables/movielens-high-loss.tex}
    \caption{Rating predictions with highest losses from MovieLens 100k.}
    \label{fig:movielens-high-loss}
\end{figure*}

\begin{figure*}
    \centering
    Spotlight 1: 3-4 star action films; high model uncertainty \\
    \input{tables/movielens-spotlight-1.tex}
    
    \vspace{1em} 
    Spotlight 2: highly rated drama films; users with few reviews \\
    \input{tables/movielens-spotlight-2}

    \vspace{1em} 
    Spotlight 3: unpopular action/comedy movies; users with many reviews 
    \input{tables/movielens-spotlight-3}
    
    \caption{Spotlights on MovieLens 100k.}
    \label{fig:movielens-spotlights}
\end{figure*}

\begin{figure*}
    \centering
    GEORGE cluster 1 \\
    \input{tables/movielens-george-1.tex}
    
    \vspace{1em} 
    
    GEORGE cluster 2 \\
    \input{tables/movielens-george-2.tex}

    \vspace{1em} 
    
    GEORGE cluster 3 \\
    \input{tables/movielens-george-3.tex}
    
    \caption{GEORGE clusters on MovieLens 100k.}
    \label{fig:movielens-george}
\end{figure*}

\begin{figure*}
    \centering
    \input{tables/squad-results}

    \caption{Spotlights on SQuAD.}
    \label{fig:squad}
\end{figure*}

%% file: tables/movielens-high-loss.tex
\rowcolors{2}{white}{gray!25}
\begin{tabular}{rrrrrrr}
\toprule
 Prediction &  Rating &  Loss &                   Movie &     Genre & Avg (\# Reviews) & User ID (\# Reviews) \\
\midrule
          4 &       1 &  11.2 &            Pulp Fiction &     Crime &        4.2 (82) &            305 (97) \\
          4 &       5 &  11.0 &     Princess Bride, The &    Action &        4.1 (58) &             419 (3) \\
          5 &       1 &  10.9 &                Face/Off &    Action &        3.9 (42) &            296 (73) \\
          4 &       1 &  10.5 &     Usual Suspects, The &     Crime &        4.3 (56) &           234 (202) \\
          4 &       1 &   8.8 &                   Fargo &     Crime &       4.3 (113) &            198 (75) \\
          3 &       5 &   8.7 &       Wizard of Oz, The & Adventure &        4.2 (46) &             358 (8) \\
          5 &       1 &   8.1 &                   Alien &    Action &        4.2 (68) &            295 (96) \\
          3 &       1 &   8.0 &                  Mother &    Comedy &        3.2 (34) &            100 (25) \\
          4 &       1 &   7.9 &               Boot, Das &    Action &        4.0 (35) &           102 (104) \\
          5 &       1 &   7.9 &    English Patient, The &     Drama &        3.7 (93) &            239 (73) \\
          5 &       1 &   7.8 &           Shallow Grave &  Thriller &        3.7 (14) &            342 (73) \\
          5 &       1 &   7.8 &                Face/Off &    Action &        3.9 (42) &           145 (131) \\
          4 &       1 &   7.7 &   Devil's Advocate, The &     Crime &        3.7 (31) &             15 (44) \\
          3 &       5 &   7.6 &    Addams Family Values &    Comedy &        3.1 (18) &            326 (74) \\
          5 &       1 &   7.5 & Raiders of the Lost Ark &    Action &        4.3 (76) &           269 (156) \\
\bottomrule
\end{tabular}

%% file: tables/movielens-spotlight-1.tex
\rowcolors{2}{white}{gray!25}
\begin{tabular}{rrrrrrr}
\toprule
 Prediction &  Rating &  Loss &                                    Movie &   Genre & Avg (\# Reviews) & User ID (\# Reviews) \\
\midrule
          4 &       2 &   1.7 &                         Romeo and Juliet &   Drama &        3.4 (27) &            13 (263) \\
          5 &       1 &   1.6 &                             Lost Highway & Mystery &        2.8 (26) &            347 (78) \\
          1 &       3 &   1.6 &                                Crow, The &  Action &        3.4 (30) &            217 (39) \\
          4 &       3 &   1.4 &                                True Lies &  Action &        3.2 (40) &            13 (263) \\
          3 &       2 &   1.8 &                                Crow, The &  Action &        3.4 (30) &            197 (66) \\
          4 &       3 &   1.7 &                            Jurassic Park &  Action &        3.6 (53) &           363 (102) \\
          5 &       4 &   1.4 &                            Happy Gilmore &  Comedy &        3.2 (19) &           145 (131) \\
          3 &       3 &   1.2 &                                Crow, The &  Action &        3.4 (30) &           109 (100) \\
          4 &       1 &   1.3 &                                    Crash &   Drama &        2.5 (35) &           286 (130) \\
          4 &       3 &   1.5 &                            Happy Gilmore &  Comedy &        3.2 (19) &            223 (53) \\
          5 &       5 &   1.0 &                      Cook the Thief, The &   Drama &        3.6 (13) &           269 (156) \\
          4 &       4 &   1.1 &                                True Lies &  Action &        3.2 (40) &            347 (78) \\
          2 &       2 &   1.1 &                         Romeo and Juliet &   Drama &        3.4 (27) &           201 (171) \\
          4 &       4 &   1.2 &                                  Gattaca &   Drama &        3.2 (23) &            13 (263) \\
          4 &       5 &   1.5 &                            Jurassic Park &  Action &        3.6 (53) &            217 (39) \\
\bottomrule
\end{tabular}

%% file: tables/movielens-spotlight-2.tex
\rowcolors{2}{white}{gray!25}
\begin{tabular}{rrrrrrr}
\toprule
 Prediction &  Rating &  Loss &                       Movie &   Genre & Avg (\# Reviews) & User ID (\# Reviews) \\
\midrule
          3 &       4 &   1.2 &                       Shine &   Drama &        4.0 (23) &            382 (20) \\
          4 &       2 &   1.9 &                   Big Night &   Drama &        4.0 (30) &            382 (20) \\
          4 &       5 &   1.4 & Madness of King George, The &   Drama &        4.0 (22) &            354 (81) \\
          5 &       3 &   1.2 &              Godfather, The &  Action &        4.4 (73) &            382 (20) \\
          4 &       3 &   1.3 &                       Bound &   Crime &        3.8 (30) &            329 (28) \\
          4 &       4 &   0.8 &                       Shine &   Drama &        4.0 (23) &            214 (65) \\
          5 &       2 &   3.4 &        Fish Called Wanda, A &  Comedy &        4.0 (50) &            370 (19) \\
          3 &       3 &   0.7 & People vs. Larry Flynt, The &   Drama &        3.6 (49) &            382 (20) \\
          4 &       2 &   3.5 &                Pulp Fiction &   Crime &        4.2 (82) &            370 (19) \\
          4 &       3 &   1.3 &         Singin' in the Rain & Musical &        4.2 (38) &            370 (19) \\
          4 &       2 &   2.4 &                       Shine &   Drama &        4.0 (23) &            116 (55) \\
          3 &       3 &   1.0 &                       Alien &  Action &        4.2 (68) &            382 (20) \\
          3 &       4 &   1.1 &                  Braveheart &  Action &        4.2 (67) &            370 (19) \\
          4 &       4 &   0.6 &                       Shine &   Drama &        4.0 (23) &            243 (36) \\
          4 &       2 &   1.9 &                       Shine &   Drama &        4.0 (23) &           201 (171) \\
\bottomrule
\end{tabular}

%% file: tables/movielens-spotlight-3.tex
\rowcolors{2}{white}{gray!25}
\begin{tabular}{rrrrrrr}
\toprule
 Prediction &  Rating &  Loss &                       Movie &     Genre & Avg (\# Reviews) & User ID (\# Reviews) \\
\midrule
          4 &       4 &   0.4 &                   Drop Zone &    Action &         2.4 (9) &           130 (175) \\
          4 &       2 &   1.8 &                  Mouse Hunt & Childrens &         2.6 (7) &             29 (17) \\
          4 &       4 &   0.9 &                Arrival, The &    Action &        2.7 (14) &           363 (102) \\
          3 &       4 &   1.1 & Father of the Bride Part II &    Comedy &        2.7 (22) &           222 (174) \\
          4 &       1 &   1.8 & Father of the Bride Part II &    Comedy &        2.7 (22) &             81 (28) \\
          4 &       3 &   1.1 &                   Drop Zone &    Action &         2.4 (9) &           393 (133) \\
          3 &       3 &   0.8 &                   Space Jam & Adventure &        2.6 (13) &           303 (208) \\
          4 &       3 &   1.3 & Father of the Bride Part II &    Comedy &        2.7 (22) &            223 (53) \\
          1 &       3 &   1.2 &                  Disclosure &     Drama &        2.7 (10) &           303 (208) \\
          4 &       3 &   1.2 &                Arrival, The &    Action &        2.7 (14) &           303 (208) \\
          3 &       3 &   0.9 &                   Space Jam & Adventure &        2.6 (13) &             21 (84) \\
          4 &       4 &   0.6 &                      Casper & Adventure &        2.6 (12) &             83 (77) \\
          4 &       2 &   1.6 &           Last Man Standing &    Action &        2.8 (14) &           303 (208) \\
          3 &       2 &   1.4 &                   Drop Zone &    Action &         2.4 (9) &            197 (66) \\
          3 &       1 &   1.6 &                Arrival, The &    Action &        2.7 (14) &           201 (171) \\
\bottomrule
\end{tabular}

%% file: tables/movielens-george-1.tex
\rowcolors{2}{white}{gray!25}
\begin{tabular}{rrrrrrr}
\toprule
 Prediction &  Rating &  Loss &                    Movie &    Genre & Avg (\# Reviews) &  User reviews \\
\midrule
          4 &       1 &  11.2 &             Pulp Fiction &    Crime &        4.2 (82) &            97 \\
          5 &       1 &  10.9 &                 Face/Off &   Action &        3.9 (42) &            73 \\
          4 &       1 &  10.5 &      Usual Suspects, The &    Crime &        4.3 (56) &           202 \\
          4 &       1 &   8.8 &                    Fargo &    Crime &       4.3 (113) &            75 \\
          5 &       1 &   8.1 &                    Alien &   Action &        4.2 (68) &            96 \\
          3 &       1 &   8.0 &                   Mother &   Comedy &        3.2 (34) &            25 \\
          4 &       1 &   7.9 &                Boot, Das &   Action &        4.0 (35) &           104 \\
          5 &       1 &   7.9 &     English Patient, The &    Drama &        3.7 (93) &            73 \\
          5 &       1 &   7.8 &            Shallow Grave & Thriller &        3.7 (14) &            73 \\
          5 &       1 &   7.8 &                 Face/Off &   Action &        3.9 (42) &           131 \\
          4 &       1 &   7.7 &    Devil's Advocate, The &    Crime &        3.7 (31) &            44 \\
          5 &       1 &   7.5 &  Raiders of the Lost Ark &   Action &        4.3 (76) &           156 \\
          4 &       1 &   7.3 &         Devil's Own, The &   Action &        2.9 (47) &           175 \\
          5 &       1 &   6.4 & Empire Strikes Back, The &   Action &        4.2 (72) &           108 \\
          4 &       1 &   6.2 &                 Clueless &   Comedy &        3.4 (24) &           121 \\
\bottomrule
\end{tabular}

%% file: tables/movielens-george-2.tex
\rowcolors{2}{white}{gray!25}
\begin{tabular}{rrrrrrr}
\toprule
 Prediction &  Rating &  Loss &                     Movie &  Genre & Avg (\# Reviews) &  User reviews \\
\midrule
          4 &       2 &   7.4 &       Grosse Pointe Blank & Comedy &        3.7 (29) &           208 \\
          5 &       2 &   7.2 &              Citizen Kane &  Drama &        4.3 (40) &           102 \\
          5 &       2 &   6.0 &                     Fargo &  Crime &       4.3 (113) &            22 \\
          4 &       2 &   5.9 &                      Jaws & Action &        3.8 (62) &            84 \\
          5 &       2 &   5.8 &          Schindler's List &  Drama &        4.4 (61) &           184 \\
          5 &       2 &   5.8 &     Sense and Sensibility &  Drama &        4.2 (51) &           131 \\
          3 &       2 &   5.6 &           Peacemaker, The & Action &        3.4 (24) &            12 \\
          4 &       2 &   5.4 &                 Star Wars & Action &        4.4 (99) &            78 \\
          5 &       2 &   5.4 &           Full Monty, The & Comedy &        4.0 (63) &           263 \\
          4 &       2 &   5.3 & Shawshank Redemption, The &  Drama &        4.5 (60) &           109 \\
          4 &       2 &   5.2 &             Groundhog Day & Comedy &        3.6 (56) &           179 \\
          4 &       2 &   5.2 &      Sweet Hereafter, The &  Drama &         3.3 (9) &            12 \\
          4 &       2 &   5.2 &       Usual Suspects, The &  Crime &        4.3 (56) &            76 \\
          4 &       2 &   5.2 &     Bullets Over Broadway & Comedy &        3.7 (24) &            78 \\
          4 &       2 &   5.2 &                   Contact &  Drama &       3.7 (107) &           149 \\
\bottomrule
\end{tabular}

%% file: tables/movielens-george-3.tex
\rowcolors{2}{white}{gray!25}
\begin{tabular}{rrrrrrr}
\toprule
 Prediction &  Rating &  Loss &                    Movie &     Genre & Avg (\# Reviews) &  User reviews \\
\midrule
          2 &       5 &   4.8 &    2001: A Space Odyssey &     Drama &        4.1 (57) &           155 \\
          4 &       5 &   4.2 &           Cool Hand Luke &    Comedy &        4.1 (31) &           143 \\
          2 &       5 &   4.2 &            Birdcage, The &    Comedy &        3.4 (62) &           217 \\
          3 &       5 &   2.8 &                   Gandhi &     Drama &        4.0 (38) &           155 \\
          3 &       5 &   2.4 &         Schindler's List &     Drama &        4.4 (61) &           155 \\
          4 &       5 &   2.2 &                Cape Fear & Film-Noir &        3.7 (20) &           155 \\
          4 &       5 &   1.7 &      Blues Brothers, The &    Action &        3.9 (45) &           155 \\
          4 &       5 &   1.5 &           Cool Hand Luke &    Comedy &        4.1 (31) &           146 \\
          4 &       5 &   1.4 &           Cool Hand Luke &    Comedy &        4.1 (31) &            98 \\
          4 &       5 &   1.3 &           Cool Hand Luke &    Comedy &        4.1 (31) &            33 \\
          4 &       5 &   1.2 &           Cool Hand Luke &    Comedy &        4.1 (31) &            47 \\
          4 &       5 &   1.1 &           Cool Hand Luke &    Comedy &        4.1 (31) &           101 \\
          5 &       5 &   1.1 &                   Grease &    Comedy &        3.6 (32) &           155 \\
          5 &       5 &   1.1 & Empire Strikes Back, The &    Action &        4.2 (72) &           155 \\
          4 &       5 &   1.0 &           Cool Hand Luke &    Comedy &        4.1 (31) &           114 \\
\bottomrule
\end{tabular}

%% file: tables/squad-results.tex
\begin{tabular}{p{0.75in} p{2.25in} p{2in}}
    \toprule
    \textbf{Subset} & \textbf{Frequent words} & \textbf{Frequent topics} \\
    \midrule
    High loss & 
    sacks, tackles, confused, yards, behavior, touchdowns, defendants, protesters, cornerback, interceptions &
    civil disobedience, 1973 oil crisis, complexity theory \\
    \midrule
    Spotlight 1 & 
    packet, packets, switching, circuit, pad, messages, dialogue, aim, networking, why, &
    packet switching, civil disobedience, computational complexity theory \\
    \midrule
    Spotlight 2 & 
    touchdowns, passes, offense, yards, receptions, rating, anderson, receiver, punt, selections &
    civil disobedience, ctenophora, yuan dynasty \\
    \midrule
    Spotlight 3 & 
    networking, alice, capacity, consequence, combining, teach, views, protest, acceleration, switching &
    packet switching, teacher, force \\
    \bottomrule
\end{tabular}